\documentclass[11pt,a4paper]{article}
\usepackage[hyperref]{emnlp2018}
\usepackage{times}
\usepackage{latexsym}
\usepackage{tabularx}
\usepackage{CJKutf8}
\usepackage{amsmath}
\usepackage[margin=2.5cm]{geometry}

\usepackage{url}
\usepackage{graphicx}
\usepackage{subcaption}
\usepackage{xcolor}
\usepackage{multirow}

\newcommand\blfootnote[1]{
  \begingroup
  \renewcommand\thefootnote{}\footnote{#1}
  \addtocounter{footnote}{-1}
  \endgroup
}

\aclfinalcopy 

\definecolor{cadmiumgreen}{rgb}{0.0, 0.42, 0.24}

\title{Character-level Chinese-English Translation through ASCII Encoding}

\author{
Nikola I. Nikolov$^{*}$, Yuhuang Hu$^{*}$, Mi Xue Tan, Richard H.R. Hahnloser \\
  Institute of Neuroinformatics, University of Z{\"u}rich and ETH Z{\"u}rich, Switzerland \\
  {\tt \{niniko, yuhuang.hu, mtan, rich\}@ini.ethz.ch}
}

\date{}

\begin{document}
\maketitle
\begin{abstract}
  Character-level Neural Machine Translation (NMT) models have recently achieved impressive results on many language pairs. They mainly do well for Indo-European language pairs, where the languages share the same writing system. However, for translating between Chinese and English, the gap between the two different writing systems poses a major challenge because of a lack of systematic correspondence between the individual linguistic units. In this paper, we enable character-level NMT for Chinese, by breaking down Chinese characters into linguistic units similar to that of Indo-European languages. We use the Wubi encoding scheme\footnote{Code and data available at \url{https://github.com/duguyue100/wmt-en2wubi}.}, which preserves the original shape and semantic information of the characters, while also being reversible. We show promising results from training Wubi-based models on the character- and subword-level with recurrent as well as convolutional models. 
  \blfootnote{* Equal contribution}

\end{abstract}

\section{Introduction}

\label{sec:intro}

\begin{CJK}{UTF8}{gbsn}

Character-level sequence-to-sequence (Seq2Seq) models for machine translation can perform comparably to subword-to-subword or subword-to-character models, when dealing with Indo-European language pairs, such as German-English or Czech-English \citep{lee2016char}. Such language pairs benefit from having a common Latin character representation, which facilitates suitable character-to-character mappings to be learned. This method, however, is more difficult for non-Latin language pairs, such as Chinese-English. Chinese characters differ from English characters, in the sense that they carry more meaning and resemble subword units in English. For example, the Chinese character `人' corresponds to the word `human' in English. This lack of correspondence makes the problem more demanding for a Chinese-English character-to-character model, as it would be forced to map higher-level linguistic units in Chinese to individual Latin characters in English. Good performance on this task may, therefore, require specific architectural decisions.
\end{CJK}

\begin{figure}[t!]
	\begin{CJK}{UTF8}{gbsn}
	\centering
    \includegraphics[width=0.30\textwidth]{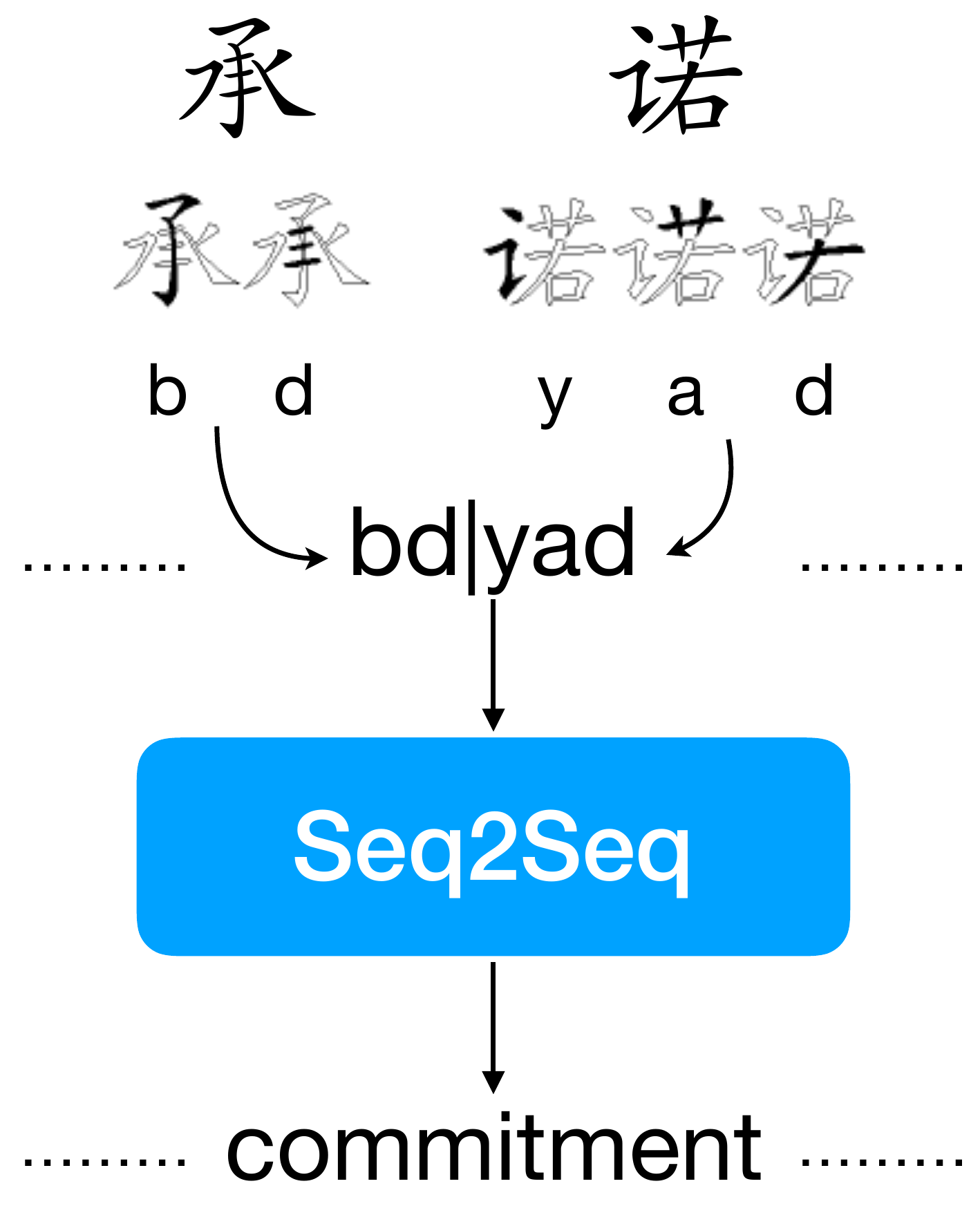}
    \caption{Overview of the \textbf{wubi2en} approach to Chinese-to-English translation. A raw Chinese word (`承诺') is encoded into ASCII characters (`bd$|$yad'), using the Wubi encoding method, before passing it to a Seq2Seq network. The network generates the English translation `commitment', processing one ASCII character at a time. 
    }\label{fig:model}
    \end{CJK}
\end{figure}

In this paper, we propose a simple solution to this challenge: encode Chinese into a meaningful string of ASCII characters, using the \textbf{Wubi} method \citep{lunde2009cjkv} (Section~\ref{sec:wubi}). This encoding enables efficient and accurate character-level prediction applications in Chinese, with no changes required to the model architecture (see Figure ~\ref{fig:model}). Our approach significantly reduces the character vocabulary size of a Chinese text, while preserving the shape and semantic information encoded in the Chinese characters. 

We demonstrate the utility of the Wubi encoding on subword- and character-level Chinese NMT, comparing the performance of systems trained on Wubi vs. raw Chinese characters (Section~\ref{sec:exps}). We test three types of Seq2Seq models: recurrent \citep{Cho2014EncDec} convolutional \citep{gehring2017CNNFB} as well as hybrid \citep{lee2016char}. Our results demonstrate the utility of Wubi as a preprocessing step for Chinese translation tasks, showing promising performance.

\section{Background}\label{sec:background}

\subsection{Sequence-to-sequence models for NMT}

Neural networks with Encoder-Decoder architectures have recently achieved impressive performance on many language pairs in Machine Translation, such as English-German and English-French \citep{wu2016google}. Recurrent Neural Networks (RNNs) \citep{Cho2014EncDec} process and \textit{encode} the input sequentially, mapping each word onto a vector representation of fixed dimensionality. The representations are used to condition a \textit{decoder} RNN which generates the output sequence. 

Recent studies have shown that Convolutional Neural Networks (CNNs)~\citep{Lecun1998CNN} can perform better on Seq2Seq tasks than RNNs~\citep{gehring2017CNNFB,chen2017cnn,Kalchbrenner2016Bytenet}. CNNs enable simultaneous computations which are more efficient especially using parallel GPU hardware. Successive layers in CNN models have an increasing receptive field for modeling long-term dependencies in candidate languages. 

\subsection{Chinese-English translation}

Recent large-scale benchmarks of RNN encoder-decoder models \citep{wu2016google,junczys2016neural} have shown that translation pairs involving Chinese are among the most challenging for NMT systems. For instance, in \citet{wu2016google} an NMT system trained on English-to-Chinese had the least relative improvement across five other language pairs, measured over the performance of a phrase-based machine translation baseline.

While it is known that the quality of a Chinese translation system can be significantly impacted by the choice of word segmentation \citep{wang2015seg}, there has been little work on improving the representation medium for Chinese translation. \citet{Wang:2017} perform an empirical comparison on various translation granularities for the Chinese-English task. They find that adding additional information about the segmentation of the Chinese characters, such
as marking the start and the end of each word, leads to improved performance over raw character or word translation. 

The work that is most related to ours is \citep{du:2017}, in which they use Pinyin\footnote{The official romanization system for Standard Chinese in mainland China.} to romanize raw Chinese characters based on their pronunciation. This method, however, adds ambiguity to the data, because many Chinese characters share the same pronunciation.

\section{Encoding Chinese characters with Wubi}\label{sec:wubi}

\begin{CJK}{UTF8}{gbsn}
\textbf{Wubi} \citep{lunde2009cjkv} is a shape-based encoding method for inputting Chinese characters on a computer QWERTY keyboard. The encoding is based on the structure of the characters rather than on their pronunciation. Using the method, each raw Chinese character (e.g., ``设'') can be efficiently mapped to a unique sequence of 1 to 5 ASCII characters (e.g., ``ymc''). This feature greatly reduces the ambiguity brought by other phonetic input methods, such as Pinyin.

As an input method, Wubi uses 25 key caps from the QWERTY keyboard, where each key cap is assigned to five categories based on the character's first stroke (when written by hand). Each of the key caps is associated with different character roots. A Chinese character is broken down into its character roots, and a corresponding QWERTY association of the character roots is used to encode a word. For example, the Wubi encoding of `哈' is `kwgk', and the character roots of this word are 口(k),　人(w),　王(g) and 口(k). To create a one-to-one mapping of every Chinese character to a Wubi encoding during translation, we append numbers to the encodings, whenever one code maps to multiple Chinese characters. 

\begin{table}[ht]
	\centering
	\caption{Examples of Wubi words and the corresponding Chinese words} 
    \begin{tabular}{c|c|c}
    	\hline
        \textbf{English} & \textbf{Chinese} & \textbf{Wubi} \\
        \hline
        Set up & 编设 & xyna0$|$ymc  \\
        Public property & 公共财产  & wc$|$aw$|$mf$|$u  \\
        Step aside & 让开 & yh$|$ga  \\
        \hline
    \end{tabular}
	\label{tab:wubiwordtable}
\end{table}

Applying Wubi significantly reduces the character-level vocabulary size of a Chinese text (from $>5,000$ commonly used Chinese characters, to $128$ ASCII characters\footnote{$302$ ASCII and special characters such as non-ASCII symbols used in the experiments, see Section~\ref{sec:exps}.}), while preserving its shape and semantic information. Table \ref{tab:wubiwordtable} contains examples of Wubi, along with the corresponding words in Chinese and English. 

\end{CJK}

\section{Results}\label{sec:exps}

\subsection{Dataset}\label{subsec:setup}

In this work, we use a subset of the English and Chinese parts of the United Nations Parallel Corpus \citep{ziemski2016UN}. We choose the UN corpus because of its high-quality, man-made translations. The dataset is sufficient for our purpose: our aim here is not to reach state-of-the-art performance on Chinese-English translation, but to demonstrate the potential of the Wubi encoding on the character level. 

\begin{CJK}{UTF8}{gbsn}
We preprocess the UN dataset with the MOSES tokenizer\footnote{\url{https://github.com/moses-smt}}, and use Jieba\footnote{\url{https://github.com/fxsjy/jieba}} to segment the Chinese sentence into words, following which we encode the texts into Wubi. We use the `$|$' character as a subword separator for Wubi, in order to ensure that the mapping from Chinese to Wubi is unique. We also convert all Chinese punctuation marks (e.g. `。、《》') from UTF-8  to ASCII (e.g. `.,$<>$') because they share similar linguistic roles to English punctuations. This conversion additionally decreases the size of the Wubi character vocabulary.
\end{CJK}

Our final dataset contains 2.1M sentence pairs for training, and 55k pairs for validation and testing respectively (Table \ref{tab:data-stat} contains additional statistics). Note that our procedures are entirely reversible.

\begin{table}[ht]
	\caption{Statistics of our dataset (mean and standard deviation).}
    \centering
    \small
    \label{tab:data-stat}
	\begin{tabular}{c|ccc}
    	\hline
        & \textbf{English} & \textbf{Wubi} & \textbf{Chinese} \\
        \hline
        words & \multirow{2}{*}{25.8$\pm$11.0} & \multirow{2}{*}{22.9$\pm$10.0} & \multirow{2}{*}{22.9$\pm$10.0} \\
        per sentence & & \\
        \hline
        characters & \multirow{2}{*}{4.9$\pm$3.3} & \multirow{2}{*}{4.6$\pm$3.3} & \multirow{2}{*}{1.8$\pm$0.83} \\
        per word & & \\
        \hline
        characters & \multirow{2}{*}{152.3$\pm$67.9} & \multirow{2}{*}{127.1$\pm$56.5} & \multirow{2}{*}{63.5$\pm$27.6} \\
        per sentence & & \\
        \hline
    \end{tabular}
\end{table}

\begin{CJK}{UTF8}{gbsn}

To investigate the utility of the Wubi encoding, we compare the performance of NMT models on four training pairs: raw Chinese-to-English (\textit{cn2en}) versus Wubi-to-English (\textit{wubi2en}); English-to-raw Chinese (\textit{en2cn}) versus English-to-Wubi (\textit{en2wubi}). For each pair, we investigate three levels of sequence granularity: word-level, subword-level, and character-level. The word-level operates on individual English words (\emph{e.g.}~walk) and either raw-Chinese words (\emph{e.g.}~编设) or Wubi words (\emph{e.g.}~sh$|$wy). We limit all word-level vocabularies to the 50k most frequent words for each language. The subword-level is produced using the byte pair encoding (BPE) scheme \citep{sennrich2015neural}, capping the vocabulary size at 10k for each language. The character-level operates on individual raw-Chinese characters (e.g. `重'), or individual ASCII characters. 
\end{CJK}

\subsection{Model descriptions and training details}

Our models are summarized in Table~\ref{tab:model-size}, including the number of parameters and vocabulary sizes used for each pair. For the subword- and word-level experiments, we use two systems\footnote{We use the \texttt{fairseq} library \url{https://github.com/pytorch/fairseq}.}. The first, \textit{LSTM}, is an LSTM Seq2Seq model~\citep{Cho2014EncDec} with an attention mechanism \citep{Bahdanau2014ONMT}. We use a single layer of 512 hidden units for the encoder and decoder, and set 512 as the embedding dimensionality. The second system, \textit{FConv}, is a smaller version of the convolutional Seq2Seq model with an attention mechanism from ~\citep{gehring2017CNNFB}. We use word embeddings with dimension 256 for this model. The encoder and the decoder of \textit{FConv} have the same convolutional architecture which consists of 4 convolution layers for the encoder and 3 for the decoder, each layer having filters with dimension 256 and size 3. 

\begin{table*}[ht]
	\caption{Model and vocabulary sizes used in our experiments. In brackets, we include the number of embedding parameters for a model (left), or the percentage of vocabulary coverage of the dataset (right).
    }
    \label{tab:model-size}
    \centering
    \small
    \begin{tabular}{c|ccc|ccc}
    	\hline
    	& \multicolumn{3}{c|}{\textbf{No. of model parameters (Embedding)}} & \multicolumn{3}{c}{\textbf{Vocab Size (\% coverage of dataset)}} \\
    	\textbf{level} & char2char & FConv & LSTM & EN & Wubi & CN \\
        \hline
        word & - & 42M (25M) & 83M (51M) & 50k (99.7\%) & 50k (99.5\%) & 50k (99.5\%) \\
        subword & - & 11M (5.1M) & 22M (10.6M) & 10k (100\%) & 10k (100\%) & 10k (98.7\%) \\
        character & 69-74M (0.21M-2.81M$^{\dagger}$) & - & - & 302 (100\%) & 302 (100\%) & 5183 (100\%) \\
        \hline
	\end{tabular}\\
    \vspace{1mm}
    $\dagger$: 0.21M for wb2en/en2wb (69M in total); 0.77M for cn2en (70M) and 2.81M for en2cn (74M), \\due to a larger size of the decoder embedding.
\end{table*}

\begin{table*}[ht]
    \caption{BLEU test scores on the UN dataset.}\label{tab:result}
    \centering
    \begin{tabular}{c|ccccc}
    	\hline
        & \textbf{character} & \multicolumn{2}{c}{\textbf{subword}} & \multicolumn{2}{c}{\textbf{word}} \\
	    & char2char & FConv & LSTM & FConv & LSTM \\
        \hline 
        wubi2en & \textbf{40.55} & 38.20 & 43.06 & 39.53 & 43.36 \\\\[-0.7em]
        cn2en & 39.60 
        &  38.20 & 43.03 & 39.64 & 43.67 \\\\[-0.7em]
        \hline
        en2wubi & \textbf{36.78} & \textbf{36.04} & \textbf{39.03} & 36.98 & 39.69 \\\\[-0.7em]
        en2cn$^{\dagger}$ & 36.13 
        & 35.41 & 38.64 & 37.25 & 39.59 \\
        \hline
    \end{tabular}\\
    \vspace{1mm}
    {\small $\dagger$: We convert these translations to Wubi before computing BLEU to ensure a consistent comparison.}
\end{table*}

For all character-level experiments, we use the fully-character level model, \textit{char2char} from \citep{lee2016char}\footnote{\url{https://github.com/nyu-dl/dl4mt-c2c}}. The encoder of this model consists of 8 convolutional layers with max pooling, which produce intermediate representations of segments of the input characters. Following this, a 4-layer highway network \citep{srivastava2015highway} is applied, as well as a single-layer recurrent network with gated recurrent units (GRUs) \citep{Cho2014EncDec}. The decoder consists of an attention mechanism and a two-layer GRU, which predicts the output one character at a time. The character embedding dimensionality is 128 for the encoder and 512 for the decoder, whereas the number of hidden units is 512 for the encoder and 1024 for the decoder. 

We train all models for 25 epochs using the Adam optimizer~\citep{Ba:2014}. We used four NVIDIA Titan X GPUs for conducting the experiments, and use beam search with beam size of 20 to generate all final outputs.

\subsection{Quantitative evaluation}

In Table~\ref{tab:result}, we present the BLEU scores for all the previously described experiments. Before computing BLEU, we convert all Chinese outputs to Wubi to ensure a consistent comparison. This conversion has a one-to-one mapping between Chinese and Wubi, whereas, in the reverse direction, ill-formed Wubi output on the character-level might not be reversible to Chinese. 

On the word-level, the Wubi-based models achieve comparable results to their counterparts in Chinese, in both translation directions. \textit{LSTM} significantly outperforms \textit{FConv} across all experiments here, most likely due to its much larger size (see Table \ref{tab:model-size}). 

On the subword-level, we observe a slight increase of about 0.5 BLEU when translating from English to Wubi instead of raw Chinese. This increase is most likely due to the difference in the BPE vocabularies: while the English and Wubi BPE rules that were learned cover 100\% of the dataset, for Chinese this is 98.7\% - the remaining 1.3\% had to be replaced by the \textit{unk} symbol under our vocabulary constraints. While the models were capable of compensating for this gap when translating to English, in the reverse direction it resulted in a loss of performance. This highlights one benefit of Wubi on the subword-level: the Latin encoding seems to give a greater flexibility for extracting suitable BPE rules. It would be interesting to repeat this comparison using much larger datasets and larger BPE vocabularies. 

Character-level translation is more difficult than word-level, since the models are expected to not only predict sentence-level semantics, but also to generate the correct spelling of each word. Our \textit{char2char} Wubi models outperformed the raw Chinese models with 0.95 BLEU points when translating to English, and 0.65 BLEU when translating from English. The differences are statistically significant ($p=0.001$ and $p=0.034$ respectively) according to bootstrap resampling \citep{koehn2004statistical} with 1500 samples. The results demonstrate the advantage of Wubi on the character-level, which outperforms raw Chinese even though it has fewer parameters dedicated for character embeddings (Table \ref{tab:model-size}) and that it has to deal with substantially longer input or output sequences (see Table \ref{tab:data-stat}). 

\begin{figure*}
    \centering
    \begin{subfigure}[b]{0.4\textwidth} \includegraphics[width=\textwidth]{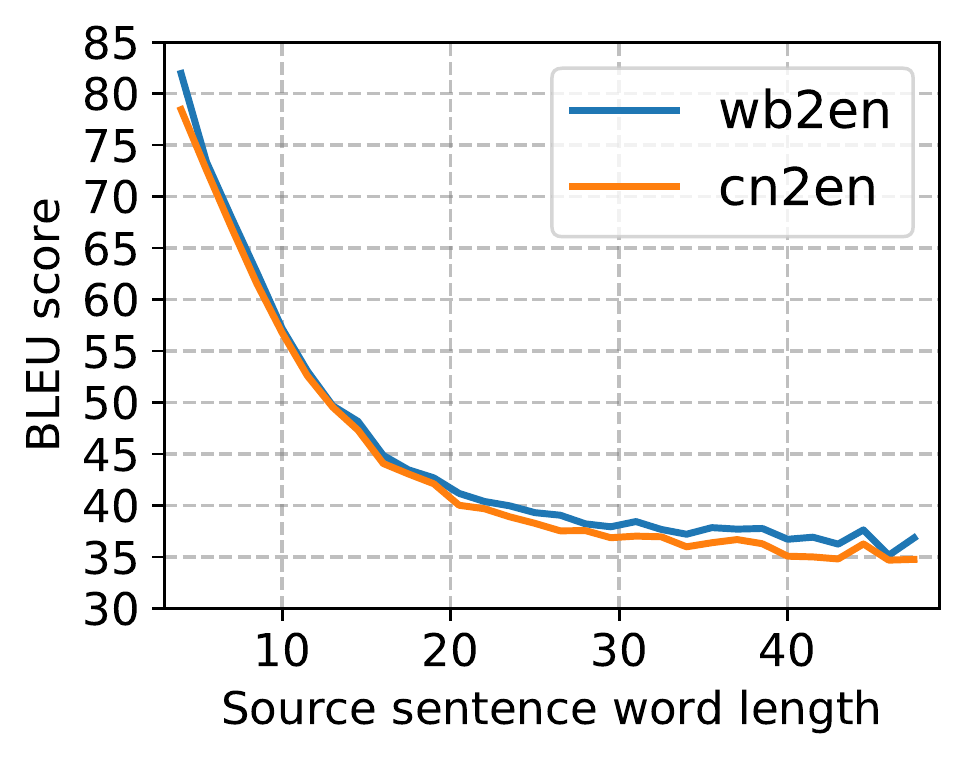}
        \caption{Translation from Chinese to English.}
        \label{fig:to_english}
    \end{subfigure}
    \begin{subfigure}[b]{0.4\textwidth} \includegraphics[width=\textwidth]{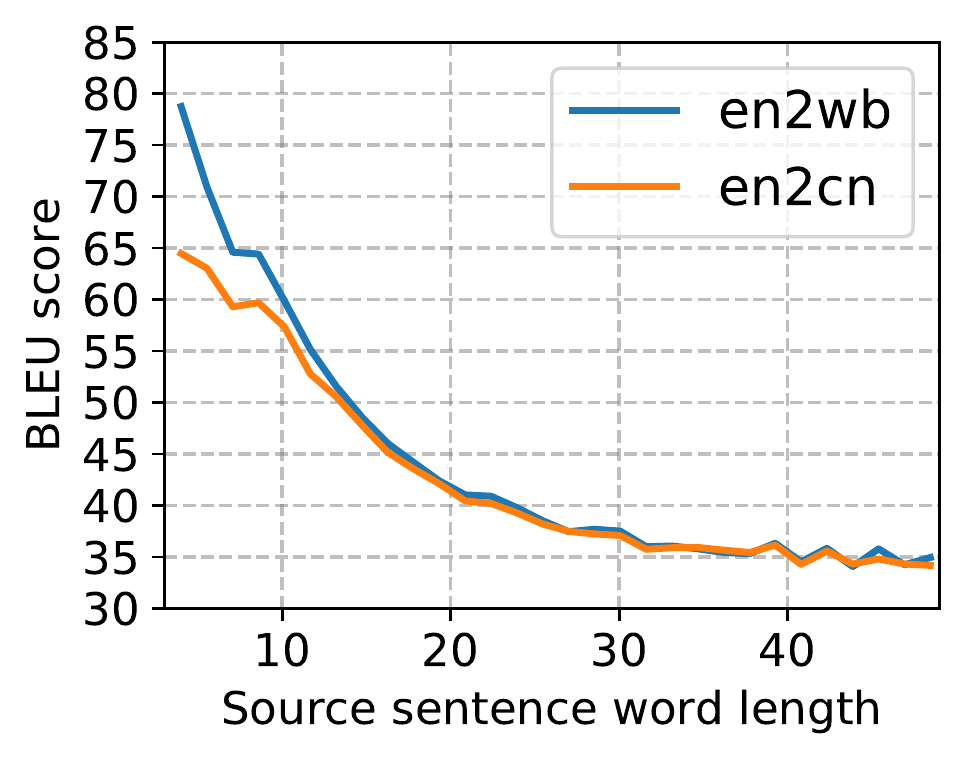}
        \caption{Translation from English to Chinese.}
        \label{fig:from_english}
    \end{subfigure}
    \caption{Sentence-level BLEU scores obtained by the character-level \textit{char2char} models on our test dataset, plotted with respect to the word length of the source sentences.}\label{fig:sent-bleu}
\end{figure*}

In Figure \ref{fig:sent-bleu}, we plot the sentence-level BLEU scores obtained by the \textit{char2char} models on our test set, with respect to the length of the input sentences. When translating from Chinese to English (Figure \ref{fig:to_english}) the Wubi-based model consistently outperforms the raw Chinese model, for all input lengths. Interestingly, the gap between the two systems increases for longer Chinese inputs of over 20 words, indicating that Wubi is more robust for such examples. This result could be explained by the fact that the encoder of the \textit{char2char} model is more suitable for modeling languages with a higher level of granularity such as English and German. When translating from English to Chinese (Figure \ref{fig:from_english}) Wubi still has a small edge, however in this case we see the reverse trend: it performs much better on shorter sentences up to 12 English words. Perhaps, the increased granularity of the output sequence led to an advantage during decoding using beam search.

\begin{table*}[!ht]
\begin{CJK}{UTF8}{gbsn}
    \caption{Four examples from our test dataset, along with system-generated translations produced by the \textit{char2char} models. We converted the Wubi translations to raw Chinese. Translations of words with a similar meaning are marked with the same color.}\label{tab:example}
\fontsize{9.36}{10}\selectfont
\begin{tabularx}{\hsize}{l|l|X}
\hline
\multicolumn{2}{c|}{\textbf{Translation Type}} & \textbf{Example 1} \\
\hline
\textbf{English} & ground truth & social \textcolor{red}{and} human rights questions \\ 
\textbf{Chinese} & ground truth & 社会 \textcolor{red}{与} 人权 问题\\
\textbf{Wubi} & ground truth & py$|$wf \textcolor{red}{gn} w$|$sc ukd0$|$jghm1$|$\\
\cline{2-3}
 & wubi2en & social \textcolor{red}{and} human rights questions\\
 & cn2en & social \textcolor{red}{and} human rights questions\\
 & en2wubi & 社会 \textcolor{red}{与} 人权 问题\\
 & en2cn & 社会 \textcolor{red}{和} 人权 问题\\
\hline
\multicolumn{2}{c|}{\ } & \textbf{Example 2} \\ \hline
\textbf{English} & ground truth & \textcolor{cadmiumgreen}{the} informal consultations \textcolor{red}{is} \textcolor{blue}{open} to all member states .\\ 
\textbf{Chinese} & ground truth & 所有 会员国 均 \textcolor{red}{可} \textcolor{blue}{参加} \textcolor{cadmiumgreen}{这次} 非正式 协商 。
\\
\textbf{Wubi} & ground truth & rn$|$e wf$|$km$|$l fqu \textcolor{red}{sk} \textcolor{blue}{cd$|$lk} \textcolor{cadmiumgreen}{p$|$uqw} djd$|$ghd0$|$aa fl$|$um .\\
\cline{2-3}
 & wubi2en & \textcolor{cadmiumgreen}{this} informal consultation \textcolor{red}{may be} \textcolor{blue}{open} to all member states .\\
 & cn2en & \textcolor{cadmiumgreen}{the} informal consultations \textcolor{red}{will be} \textcolor{blue}{open} to all member states .\\
 & en2wubi & 所有 会员国 均 \textcolor{red}{可} \textcolor{blue}{参加} \textcolor{cadmiumgreen}{这次} 非正式 协商 。\\
 & en2cn & 所有 会员国 均 \textcolor{red}{可} \textcolor{blue}{进行} 非正式 协商 。\\
\hline
\multicolumn{2}{c|}{\ } & \textbf{Example 3} \\
\hline
\textbf{English} & ground truth &  we \textcolor{blue}{believe} that increased trade is \textcolor{red}{essential} for the growth and development of \textcolor{cadmiumgreen}{ldcs} .\\ 
\textbf{Chinese} & ground truth & 我们 \textcolor{blue}{相信} ， 增加 贸易 对 \textcolor{cadmiumgreen}{最 不 发达国家} 的 增长 和 发展 \textcolor{red}{至 为 重要} 。\\
\textbf{Wubi} & ground truth & q$|$wu \textcolor{blue}{sh$|$wy} , fu$|$lk qyv$|$jqr cf \textcolor{cadmiumgreen}{jb i v$|$dp$|$l$|$pe} r fu$|$ta t v$|$nae \textcolor{red}{gcf o tgj$|$s} .\\
\cline{2-3}
 & wubi2en & we \textcolor{blue}{believe} that increased trade is \textcolor{red}{essential} for the growth and development of \textcolor{cadmiumgreen}{the least developed countries} .\\
 & cn2en & we \textcolor{blue}{are convinced} that increased trade growth and development is \textcolor{red}{essential} .\\
 & en2wubi & 我们 \textcolor{blue}{认为} ， 增加 贸易 对 \textcolor{cadmiumgreen}{最 不 发达国家} 的 增长 和 发展 \textcolor{red}{至关重要} 。\\
 & en2cn & 我们 \textcolor{blue}{认为} ， 增加 贸易 对于 \textcolor{cadmiumgreen}{最 不 发达国家} 的 增长 和 发展 来说 是 \textcolor{red}{必不可少} 的 。\\
\hline
\multicolumn{2}{c|}{\ } & \textbf{Example 4} \\ \hline
\textbf{English} & ground truth & in some cases , additional posts were \textcolor{blue}{requested} \textcolor{red}{without explanation} .\\ 
\textbf{Chinese} & ground truth & 在 某些 情况 中 ， \textcolor{blue}{提出} 增加 员额 \textcolor{cadmiumgreen}{要求} 时 ， \textcolor{red}{并未 作出 说明} 。\\
\textbf{Wubi} & ground truth & d afs$|$hxf nge$|$ukq k , \textcolor{blue}{rj$|$bm} fu$|$lk km$|$ptkm0 \textcolor{cadmiumgreen}{s$|$fiy} jf , ua$|$fii wt$|$bm yu$|$je .\\
\cline{2-3}
 & wubi2en & in some cases , \textcolor{red}{no indication was made} when additional staffing \textcolor{cadmiumgreen}{requirements} were \textcolor{blue}{proposed} .\\
 & cn2en & in some cases , there was \textcolor{red}{no indication} of \textcolor{cadmiumgreen}{the request} for additional posts .\\
 & en2wubi & 在 有些 情况 下 ， \textcolor{cadmiumgreen}{要求} 增加 员额 。\\
 & en2cn & 在 有些 情况 下 还 \textcolor{cadmiumgreen}{要求} 增设 员额 ， 但 \textcolor{red}{没有 作出 任何 解释} 。\\
\hline
\end{tabularx}
\end{CJK}
\end{table*}

Interestingly, all the \textit{char2char} models use only a tiny fraction of their parameters as embeddings, due to the much smaller size of their vocabularies. The best-performing LSTM word-level model has the majority of its parameters, 61\% or over 50M, dedicated to word embeddings. For the Wubi-based character-level models, the number is only 0.3\% or 0.21M. There is even a significant difference between Wubi and Chinese on the character-level, for example, \textit{en2wb} has 12 times fewer embedding parameters than \textit{en2cn}. Thus, although \textit{char2char} performed worse than \textit{LSTM} in our experiments, these results highlight the potential of character-level prediction for developing compact yet performant translation systems, for Latin as well as non-Latin languages.

\subsection{Qualitative evaluation}

\begin{CJK}{UTF8}{gbsn}

In Table~\ref{tab:example}, we present four examples from our test dataset that cover short as well as long sentences. We also include the translations produced by the character-level \textit{char2char} systems, which is the main focus of this paper. Full examples from the additional systems are available in the supplementary material.

In the first example, which is a short sentence resembling the headline of a document, both the \textit{wubi2en} and {cn2en} models produced correct translations. When translating from English to Chinese, however, the \textit{en2wubi} produced the word `与' (highlighted in red) which more correctly matches the ground truth text. In contrast, the \textit{en2cn} model produced the synonym `和'. In the second example, the \textit{en2wubi} output completely matches the ground truth and is superior to the \textit{en2cn} output. The latter failed to correctly translate `the' to `这次' (marked in green).

The \textit{wubi2en} translation in the third example accurately translated the word `believe' (marked in blue) and the full form of the abbreviation `ldcs' -- `the least developed countries' (highlighted in green), whereas the \textit{cn2en} chooses `are convinced' and ignores `ldcs' in its output sentence. Interestingly, although the ground truth text maps the word `essential' (marked in red) to three Chinese words `至\textvisiblespace 为\textvisiblespace 重要', both \textit{en2wubi} and \textit{en2cn} use only a single word to interpret it. Arguably, \textit{en2wubi}'s translation `至关重要' is closer to the ground truth than \textit{en2cn}'s translation `必不可少'.

The fourth example is more challenging. There, the English ground truth `requested' (highlighted in blue) maps to two different parts of the Chinese ground truth `提出' (in blue) and `要求' (in green).
This one-to-many mapping confuses both translation models. The \textit{wubi2en} tries to match the Chinese text by translating `提出' into `proposed' and `要求' into `requirements': this model may have been misled by the word `时' (can be translated to `when'); the output contains an adverbial clause. While the \textit{wubi2en} output is closer to the ground truth, the two have little overlap. For the English-to-Chinese task, the \textit{en2cn} translation is better than the one produced by \textit{en2wubi}: while \textit{en2cn} successfully translated `without explanation' (in red), the \textit{en2wubi} model ignored this part of the sentence.

The Wubi-based models tend to produce slightly shorter translations for both directions (see Table~\ref{tab:pred:len}). In overall, the Wubi-based outputs appear to be visibly better than the raw Chinese-based outputs, in both directions. 

\begin{table}[ht]
	\centering
    \caption{Word counts of the outputs of the \textit{char2char} models (mean and standard deviation).}\label{tab:pred:len}
    \begin{tabular}{ll}
    	\hline
        \textbf{Model} & \textbf{Word Count} \\
        \hline
        \textbf{wb2en} & $25.01 \pm 10.95$ \\
        \textbf{cn2en} & $25.80 \pm 11.72$ \\
        \hline
        \textbf{en2wb} & $21.61 \pm 9.68$ \\
        \textbf{en2cn} & $22.19 \pm 10.11$ \\
        \hline
    \end{tabular}
\end{table}
\end{CJK}

\section{Conclusion}\label{sec:conclusion}

We demonstrated that an intermediate encoding step to ASCII characters is suitable for the character-level Chinese-English translation task, and can even lead to performance improvements. All of our models trained using the Wubi encoding achieve comparable or better performance to the baselines trained directly on raw Chinese. On the character-level, using Wubi yields BLEU improvements when translating both to and from English, despite the increased length of the input or output sequences, and the smaller number of embedding parameters used. Furthermore, there are also improvements on the subword-level, when translating from English. 

Future work will focus on making use of the semantic structure of the Wubi encoding scheme, to develop architectures tailored to utilize it. Another exciting future direction is multilingual many-to-one character-level translation from Chinese and several Latin languages simultaneously, which becomes possible using encodings such as Wubi. This has previously been successfully realized for Latin and Cyrillic languages \citep{lee2016char}. 

\section*{Acknowledgments}

We acknowledge support from the Swiss National Science Foundation (grant 31003A\_156976) and from the National Centre of Competence in Research (NCCR) Robotics.


\end{document}